\documentclass[a4paper,11pt]{article} 

\usepackage{authblk} 
\usepackage[utf8]{inputenc}
\usepackage[english]{babel}
\usepackage{csquotes} 
\usepackage{graphicx}
\usepackage{amsmath,amssymb,amsfonts}
\usepackage{amsthm}
\usepackage{mathrsfs}
\usepackage{xcolor}
\usepackage{textcomp}
\usepackage{booktabs}
\usepackage{rotating}
\usepackage{makecell} 
\usepackage[title]{appendix}
\usepackage{setspace}
\usepackage[left]{lineno} 

\usepackage[authoryear,round]{natbib}

\usepackage[margin=3cm]{geometry}
\usepackage[hidelinks]{hyperref}

\begin{document}

\title{\textbf{Criteria-first, semantics-later: reproducible structure discovery in image-based sciences}}

\renewcommand\Authfont{\Large}      
\renewcommand\Affilfont{\normalsize}    
\author[1,2,3]{Jan Bumberger%
  \thanks{ORCID: \href{https://orcid.org/0000-0003-3780-8663}{\nolinkurl{0000-0003-3780-8663}}}
}
\affil[1]{Helmholtz Centre for Environmental Research -- UFZ, Research Data Management - RDM, Permoserstraße 15, Leipzig, 04318, Germany}
\affil[2]{Helmholtz Centre for Environmental Research -- UFZ, Department Monitoring and Exploration Technologies, Permoserstraße 15, Leipzig, 04318, Germany}
\affil[3]{German Centre for Integrative Biodiversity Research (iDiv) Halle-Jena-Leipzig, Puschstraße 4, Leipzig, 04103, Germany}

\date{} 

\maketitle


\begin{abstract}
Across the natural and life sciences, images have become a primary measurement modality, yet the dominant analytic paradigm remains semantics-first. Structure is recovered by predicting or enforcing domain-specific labels. This paradigm fails systematically under the conditions that make image-based science most valuable, including open-ended scientific discovery, cross-sensor and cross-site comparability, and long-term monitoring in which domain ontologies and associated label sets drift culturally, institutionally, and ecologically. A deductive inversion is proposed in the form of criteria-first and semantics-later. A unified framework for criteria-first structure discovery is introduced. It separates criterion-defined, semantics-free structure extraction from downstream semantic mapping into domain ontologies or vocabularies and provides a domain-general scaffold for reproducible analysis across image-based sciences. Reproducible science requires that the first analytic layer perform criterion-driven, semantics-free structure discovery, yielding stable partitions, structural fields, or hierarchies defined by explicit optimality criteria rather than local domain ontologies. Semantics is not discarded; it is relocated downstream as an explicit mapping from the discovered structural product to a domain ontology or vocabulary, enabling plural interpretations and explicit crosswalks without rewriting upstream extraction. Grounded in cybernetics, observation-as-distinction, and information theory’s separation of information from meaning, the argument is supported by cross-domain evidence showing that criteria-first components recur whenever labels do not scale. Finally, consequences are outlined for validation beyond class accuracy and for treating structural products as FAIR, AI-ready digital objects for long-term monitoring and digital twins.
\end{abstract}


\section{Why semantics-first is now the limiting assumption}
\label{sec:why-semantics-first-limiting-assumption}

Images are a primary measurement modality across the natural and life sciences, yet analysis still often defaults to a semantics-first paradigm. In this view, structure in data is characterised by mapping measurements to a predefined domain ontology, with label sets as one common special case such as classes, object types, land-cover categories, or phenotypes. This ontology-centric framing breaks down under the very conditions that make image-based science most powerful, including long-term monitoring, cross-sensor and multi-site variability, and open-ended scientific discovery. In such settings, domain ontologies and their associated label sets drift over time and between scientific communities \citep{Kuhn1962,Bowker1999}. These are precisely the conditions targeted by monitoring agendas and digital-twin approaches, which require stable, comparable digital-twin state variables over long horizons even as interpretive schemes evolve \citep{Rossmann2022,Hazeleger2024,NASEM2024DigitalTwins}.\\

Nevertheless, across image-based sciences the default framing remains semantic. Measurements are used to predict a predefined domain ontology or label set, typically via supervised or weakly supervised pipelines whose success is evaluated primarily by agreement with that ontology. This orientation is increasingly visible in domain reviews, benchmarks, and segmentation pipelines, for example in remote sensing and medical imaging \citep{Lv2023,Xu2024}. It is further reinforced by the rise of foundation models for segmentation and vision \citep{Rodrigues2024,Zhou2024}. While such models can provide impressive generic mask proposals (e.g., SAM; \citealp{Kirillov2023}), they are often deployed as label amplifiers. Large-scale pretraining on curated corpora and/or substantial labelled supervision for adaptation and alignment, together with prompt engineering, are used to produce stable, domain-aligned outputs. This practice amplifies ontology-centric evaluation cultures and annotation demand. At the same time, these models can also be read as pre-semantic structural extractors. They can generate mask proposals or dense feature fields (e.g., DINOv2; \citealp{Oquab2024}), which can instantiate candidate structural products upstream under explicit criteria.\\

The practical success of semantic labelling can obscure a foundational problem: semantics is not a property of the image; it is a property of a community’s interpretive scheme. Ontologies and labels are negotiated, historically contingent, and purpose-oriented \citep{Bowker1999}. In practice, land-cover ontologies differ by institution and policy regime, phenotypes and cell types are reorganised as assays change, and morphological catalogues are revised with new surveys and processing pipelines. These challenges are consequential and they obstruct three core scientific aims:
(i) comparability and monitoring (long-term inference requires stability of the analytic layer even as the domain vocabulary evolves), (ii) domain shift (sensor, illumination, seasonality, or site changes), and (iii) open-ended scientific discovery (new phenomena not represented in the training label space).\\

Under these conditions, semantics-first pipelines quietly conflate two different operations: (i) recovering a structural product from measurement, and (ii) assigning meaning to that product by mapping it into a domain ontology. When meaning is imposed too early, upstream structure becomes hostage to particular domain ontologies, reducing transferability and undermining reproducibility. What is needed is not less theory, but a better placement for theory. Theory should enter as explicit criteria (optimality, stability, scale coherence), not as implicit labels. The proposed shift is therefore to relocate semantics downstream. A reproducible, criterion-defined structural layer can remain stable under declared perturbations and scale changes, while multiple semantic mappings (and evolving domain ontologies) can be applied and revised without rewriting the foundational measurement-to-structure step.\\

Classification systems are not merely descriptive; they function as epistemic infrastructures maintained through standards, documentation, and work practices, and they drift as those practices drift \citep{Bowker1999}. In Kuhnian terms, categories stabilise during phases of \enquote{normal science} and are periodically reorganised when anomalies accumulate \citep{Kuhn1962}. Semantics-first pipelines bake these contingent ontology choices into the analytic layer, turning evolving interpretive schemes into upstream constraints. A criteria-first approach does not eliminate interpretation; it decouples interpretive drift from measurement-to-structure operations by relying on a criterion-defined structural layer, so that scientific comparison remains possible even when label sets and domain ontologies evolve.

\section{Approach: criteria-first, semantics-later}
\label{sec:deductive-inversion-criteria-first-semantics-later}

A criteria-first approach is proposed, in which analysis proceeds by making explicit criteria first and addresses semantics only in a second step (semantics later). The first analytic layer derives a structural product from raw measurements using explicit optimality criteria, yielding a semantics-free structural product (e.g., partitions, graphs, fields including scalar trait fields, or hierarchies) defined by information in the measurement stream rather than by any domain ontology. Because its construction is fully specified by declared criteria, the resulting structural product is reproducible by design and transferable across domains. This separation treats the criterion-defined structural layer as a durable digital artefact for long-term monitoring, downstream reuse, and open-ended scientific discovery.

\paragraph{Key principle}
In image-based science, reproducible analysis requires that structure precedes semantics: structure is defined as a property of the measurement stream under explicit criteria, whereas semantics is a community- and purpose-bound mapping applied downstream into a domain ontology. Criterion-driven structure discovery is therefore the reproducible basis for interpretation, domain transfer, long-term monitoring, and open-ended scientific discovery.\\

\begin{figure}[t]
  \centering
  \includegraphics[width=\linewidth]{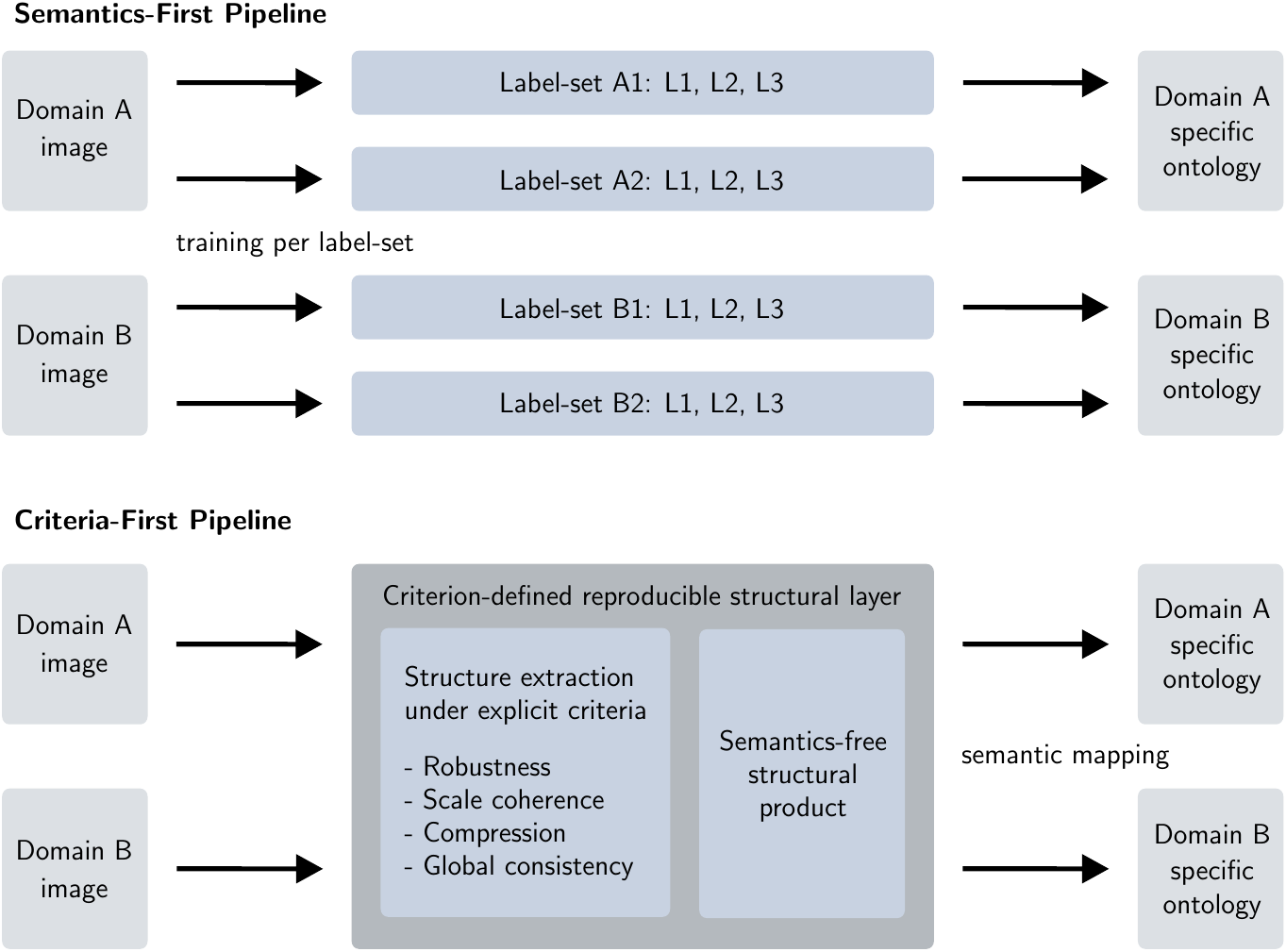}
	\caption{\textbf{The inversion.} Top: a semantics-first pipeline in which a domain-specific label set determines model training (features $\rightarrow$ prediction) and yields outputs tied to a domain ontology -- typically brittle under domain shift. Bottom: a criteria-first pipeline in which explicit optimality criteria define a reproducible, semantics-free structural product that can be mapped downstream to multiple domain ontologies (and evolving label sets).}
  \label{fig:inversion}
\end{figure}

Nevertheless, semantics-free does not mean theory-free, because every analytic procedure commits to assumptions; the difference is where and how assumptions enter. In the proposed inversion, assumptions enter upstream as explicit, inspectable criteria (e.g., stability and scale coherence; \citealp{Witkin1984,Koenderink1984,Lindeberg2008}; global optimisation objectives; \citealp{Shi2000}; trade-offs between fidelity and regularity in variational formulations; \citealp{Mumford1989}). The aim is not to remove semantics but to separate two layers: upstream, derive a structural product that is stable under declared criteria and perturbations; downstream, map that product into semantic interpretation via a domain ontology, acknowledging that this mapping is purpose- and community-dependent. Multiple semantic mappings may legitimately coexist and evolve; documenting them explicitly makes interpretive commitments auditable rather than implicit (cf.\ documentation practices such as datasheets; \citealp{Gebru2021}). Key terms are defined in Box 1, and the proposed inversion from semantics-first pipelines to criteria-first structure extraction (with semantics applied downstream) is summarised in Fig.~\ref{fig:inversion}.\\

\bigskip
\noindent
{\setlength{\fboxsep}{0.6\baselineskip}
 \setlength{\fboxrule}{0.6pt}
\fbox{%
  \begin{minipage}{0.98\linewidth}
  \small
  \textbf{Box | Definitions}\par
  \vspace{0.6\baselineskip}
  \textbf{Semantics-first:} analytic pipelines where the primary objective is mapping measurements directly to a predefined domain ontology.\par
  \vspace{0.6\baselineskip}
  \textbf{Criteria-first:} analytic pipelines where the primary objective is extracting a structural product under explicit optimality and stability criteria, independent of any domain ontology.\par
  \vspace{0.6\baselineskip}
  \textbf{Semantics-free structural product:} partitions, graphs, fields, or hierarchies defined from the measurement stream without reference to class labels or ontology terms; intended to be stable and transferable.\par
  \vspace{0.6\baselineskip}
  \textbf{Downstream semantics:} semantic mappings from a structural product into a domain ontology (including specific label sets), explicitly documented and purpose-bound.
  \end{minipage}%
}}

\section{From measurement to meaning}
\label{sec:from-measurement-to-meaning}

The proposed approach is not merely a pragmatic response to label scarcity. It follows from a deeper epistemic perspective's central claim. Meaning is not an intrinsic property of an image, but an outcome of coupling between an observing system and the measured world. What can be made reproducible across observers, disciplines, and decades is therefore not the meaning of the image, but the structural product that can be extracted from the measurement stream under explicit, shareable criteria. A classical lesson from early vision aligns with a least-commitment design principle. It recommends postponing irreversible semantic commitments and instead computing stable intermediate descriptions \citep{Marr1976,Marr1980}.\\

In a cybernetic framing, measurement is communication. A sensor delivers a message about an underlying process through a channel with a transfer function and noise \citep{Wiener1948,Ashby1956,Yuille2010}. Image analysis then begins as decoding under constraints -- recovering stable distinctions that the channel carries and that support prediction, comparison, and feedback. Crucially, those constraints need not be semantic, because they can be stated as explicit criteria such as stability under perturbations, coherence across scale, bounded complexity, or global consistency. System-theoretic accounts sharpen this boundary: observation is not passive copying but an operation of drawing distinctions; semantics is the interpretive scheme that makes distinctions communicable within a community, and that scheme is contingent and revisable \citep{Bertalanffy1968,SpencerBrown1969,Maturana1970,Bateson1972,Foerster1981}. The methodological consequence is clear: reproducible analysis must begin with distinctions definable under explicit criteria, before any community-specific interpretive scheme is imposed.\\

In terms of information theory, Shannon’s classical separation of information from meaning treats communication as uncertainty reduction under constraints. This provides a principled motivation for criterion-defined structure \citep{Shannon1959}. From this viewpoint, many structure-extraction methods are transparent as explicit criteria: thresholding maximises a separation criterion \citep{Otsu1979}; scale-space formalises structure as what persists across scales \citep{Witkin1984,Koenderink1984,Lindeberg2008}; variational formulations trade data fidelity against boundary complexity \citep{Mumford1989}; and graph partitions optimise global cut objectives \citep{Shi2000}. These are transportable optimality principles \citep{FloresFuentes2024}. Across these traditions, the same principle recurs. Semantics-free does not mean theory-free. Theory should enter upstream as explicit, inspectable criteria and declared stability commitments, not as implicit semantic assumptions. This makes structure discovery reproducible and supports domain transfer. A stable, criterion-defined structural layer can be mapped to different domain ontologies across time and across communities.


\section{Unifying framework for criterion-defined structural discovery}
\label{sec:unifying-scaffold}

\begin{figure}[htb!]
  \centering
  \includegraphics[width=\linewidth]{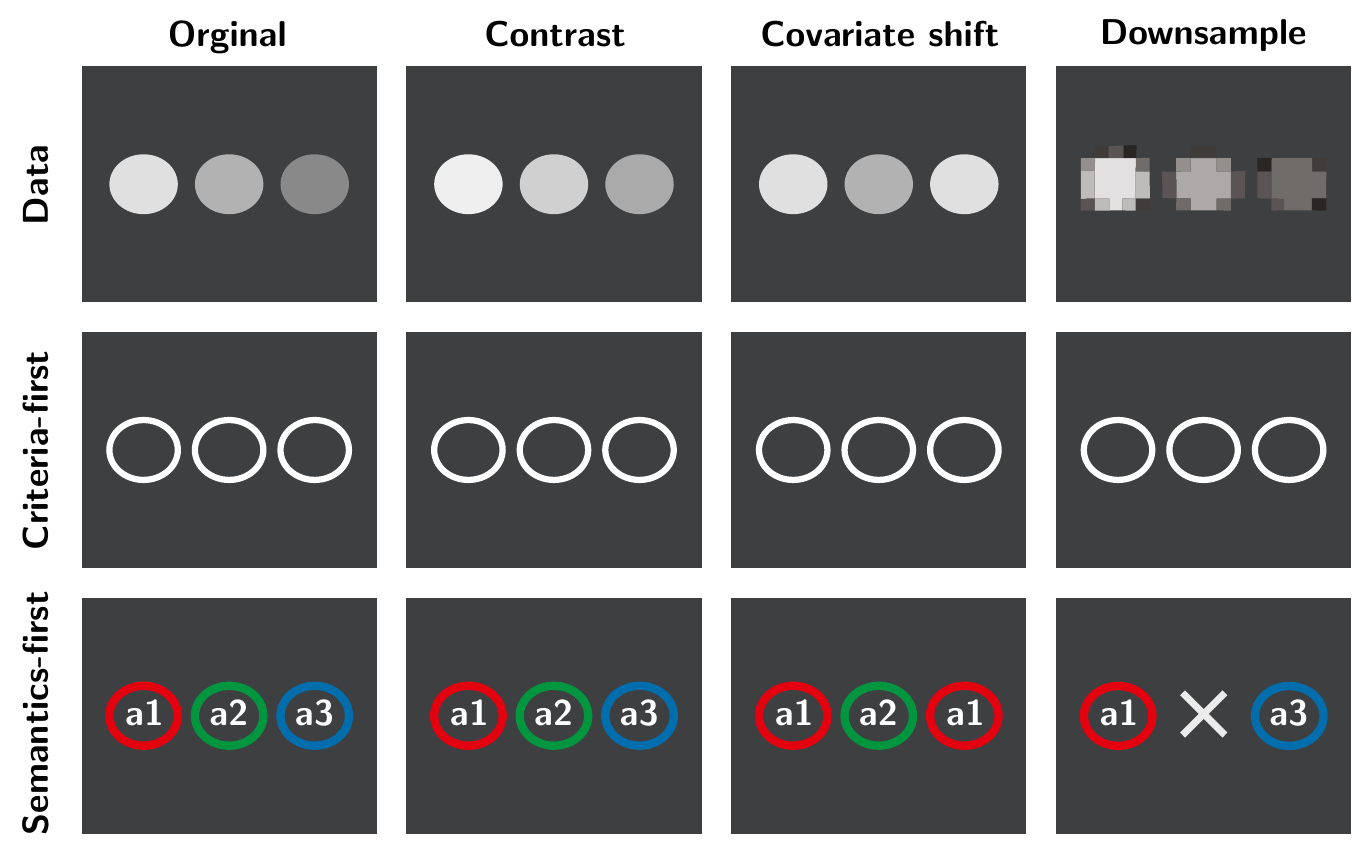}
\caption{\textbf{One image, two layers: stable structural product \(S=S_C(X)\) versus brittle semantics-first labelling under shift.}
Columns show the original synthetic measurement field \(X\) and three perturbations: global contrast change, covariate shift in appearance, and downsampling.
Top row: \(X\).
Middle row (criteria-first): \(S=S_C(X)\) under the same fixed criterion \(C\), yielding comparable object instances/boundaries across perturbations (white outlines).
Bottom row (semantics-first): labels are predicted directly from \(X\) into a fixed label set (three colour-coded labels); assignments can collapse under covariate shift or disappear under downsampling (×).
In a semantics-later framing, interpretation is a revisable mapping \(M_i:S\rightarrow \mathcal{O}_i\), so ontology drift changes \(M_i\) while \(S\) can remain comparable and structurally validated.}
\label{fig:one-image-two-criteria}
\end{figure}

Across domains, intermediate entities differ in name and domain ontology, but the analytic pattern is the same. A measurement field is transformed into a transferable structural product under explicit criteria. To keep the argument domain-general, a minimal formal setup is used, consisting of a measurement field \(X\), an explicit criterion \(C\), and a criterion-parameterised structure-extraction operator \(S_C\) whose output is a structural product \(S\). Consider an image-based measurement as a (possibly multi-channel) field
\begin{equation}
  X:\Omega \rightarrow \mathbb{R}^k,
\end{equation}
where \(\Omega\) is the carrier set (pixels/voxels, points, a spatio-temporal grid, or any sampled domain) and \(k\) denotes the number of channels. In long-term monitoring, one may equivalently view \(X\) as a measurement stream indexed over time; the definitions below apply pointwise or over spatio-temporal \(\Omega\). Here, \(C\) specifies how distinctions are drawn operationally from \(X\). \(C\) is treated as a fully specified, inspectable object (including parameters and implementation), encoding requirements such as homogeneity/contrast, boundary evidence/discontinuity, geometric consistency, topological persistence, stability under perturbations, bounded complexity/compressibility, or cross-scale coherence. Formally, \(C\) may be represented as a functional, a family of constraints, or an energy \(E_C\) defined over admissible structural candidates. Given \(X\) and \(C\), a criterion-parameterised structure-extraction operator yields the structural product
\begin{equation}
  S = S_C(X),
\end{equation}
where \(S_C\) denotes a fully specified and inspectable structure-extraction procedure parameterised by the explicit criterion \(C\). Its specification includes constraints, parameters, implementation details (including software version), and an executable protocol. The resulting structural product \(S\) is intended to be reproducible under declared stability commitments. Crucially, \(C\) does not prescribe a domain ontology (or label set); it specifies an operational notion of optimal structure under \(C\). Many concrete methods implement \(S_C\) by solving a criterion-defined decision problem. Where criteria admit convex relaxations, solutions can become effectively independent of initialization, strengthening determinacy and repeatability claims \citep{Pock2009}. A common form is energy minimisation or constrained optimisation: \begin{equation} \widehat{S} \in \operatorname*{arg\,min}_{S\,\in\, \mathcal{S}_{\mathrm{cand}}} E_C(X,\,S), \end{equation} where \(E_C(X,\,S)\) is induced by \(C\) (data fidelity + regularisation, graph-cut objectives, description length, persistence/stability penalties, etc.), and \(\mathcal{S}_{\mathrm{cand}}\) denotes the admissible candidate set. The approach deliberately does not fix a single structural type. Depending on modality and criterion, \(S\) takes values in different structural spaces. Here \(\mathcal{P}(\Omega)\) denotes the space of admissible partitions of \(\Omega\); \(\mathcal{G}\), \(\mathcal{H}\), and \(\mathcal{F}\) denote admissible spaces of graphs, hierarchies, and structural fields, respectively. For example: 
\begin{itemize} 
\item partition \(\Pi=\{A_i\}_{i=1}^n\) of \(\Omega\), with \(\Pi \,\in\, \mathcal{P}(\Omega)\); 
\item graph \(G=(V,E)\), with \(G \,\in\, \mathcal{G}\) (e.g., adjacency/contact/pose graphs); 
\item hierarchy \(H \,\in\, \mathcal{H}\) (merge trees, dendrograms, nested partitions); 
\item structural field \(F \,\in\, \mathcal{F}\) (e.g., level sets, orientation/coherence fields, scalar trait fields, continuous-valued structure descriptors). 
\end{itemize} 
The deductive commonality is that a reproducible structural product is defined as the output of explicit criteria applied to measurement. Many structural products admit a natural ordering by refinement/coarsening. For partitions, write \(\Pi^{(i)} \preceq \Pi^{(j)}\) to denote that \(\Pi^{(j)}\) is coarser than \(\Pi^{(i)}\) (each cell of \(\Pi^{(j)}\) is a union of cells of \(\Pi^{(i)}\)). A multiscale representation can then be written as a chain (or, more generally, a partially ordered family)
\begin{equation}
  \Pi^{(0)} \preceq \Pi^{(1)} \preceq \dots \preceq \Pi^{(m)}.
\end{equation}
Here \(\Pi^{(0)}\) corresponds to the sampling induced by \(\Omega\), while subsequent levels are criterion-defined aggregations. Semantic interpretation is introduced after structure discovery as a semantic mapping from the structural product (or a chosen level of description) into a domain ontology/vocabulary:
\begin{equation}
  M_i: S \rightarrow \mathcal{O}_i.
\end{equation}
Here \(\mathcal{O}_i\) denotes a domain ontology (with a label set as a common special case). Operationally, \(M_i\) often acts on a selected level \(\Pi^{(j)}\) within a multiscale family, but this choice is task- and community-dependent. Multiple mappings \(M_i:S\rightarrow \mathcal{O}_i\) may legitimately coexist for the same structural product (pluralism), reflecting different purposes, communities, or reporting regimes. This makes the key deductive point operational: reproducibility and transfer reside upstream in explicit criteria and stable structure extraction \((C,S_C)\), while domain ontologies remain downstream, purpose-bound, and revisable. Fig.~\ref{fig:one-image-two-criteria} illustrates this separation in a minimal example.\\

For a deductive reading, four postulates define the criterion-defined structural layer as a reproducibility substrate:
(i) Explicitness: the criterion \(C\) is fully specified (including parameters, software version, protocol);
(ii) Determinacy: for a given \((X,C)\), the operator \(S_C\) yields a well-defined structural product \(S\);
(iii) Stability: salient properties of \(S\) are stable under a declared family of perturbations/scale changes; and
(iv) Mapping pluralism: multiple semantic mappings \(M_i:S\rightarrow \mathcal{O}_i\) may legitimately coexist for the same \(S\).
These postulates place implicit \enquote{theory} where it can be inspected and transported -- in \(C\) and in testable stability commitments -- rather than in implicit label systems.

\section{Cross-Domain Evidence}
\label{sec:cross-domain-evidence}

Across image-based sciences, semantics-first framing remains the dominant default and is widely reflected in review and benchmark cultures \citep{Lv2023,Bom2023,Xu2024,Sun2024,Monteiro2024,Mueller2024,Yan2025}. Yet across multiple disciplines, a criteria-first approach reliably emerges whenever semantic labelling is scarce, unstable, contested, or prohibitively expensive. The recurring pattern is a practical separation in which a structural product \(S\) is extracted first under explicit criteria, while semantic interpretation is applied later as a purpose- and community-dependent mapping into a domain ontology/vocabulary. Table~\ref{tab:crossdomain_scaffold} summarises this correspondence in a domain-general scaffold: carrier sets \(\Omega\), typical operator/solver types realising \(S_C\), structural-product types, and recurring families of criteria \(C\). Extended domain overviews and respective references are provided in Supplementary Note~\ref{app:supplementary-note-domain-sketches}.

\begin{sidewaystable}[p] 
\centering
\small
\setlength{\tabcolsep}{4pt}
\renewcommand{\arraystretch}{1.15}

\begin{tabular}{p{2.6cm} p{2.2cm} p{2.7cm} p{2.5cm} p{4.8cm} p{3.8cm}}
\toprule
\textbf{Domain} &
\textbf{\(\Omega\)} &
\textbf{Operator/ \newline solver type} &
\textbf{Structural-product type} &
\textbf{Typical criteria families \(C\)} &
\textbf{Example practice terms}\\
\midrule
Earth observation (EO) and environmental monitoring&
pixels / \newline spatio-temporal grid &
segmentation; region adjacency; scale-space; criteria-driven aggregation &
partition/\allowbreak graph/\allowbreak field &
spectral homogeneity; heterogeneity; scale coherence; robustness &
pixel \(\rightarrow\) segment \(\rightarrow\) zone/patch \(\rightarrow\) region\\

Medical imaging &
voxels/slices &
variational; graph-cut; \newline level sets &
partition/\allowbreak field/\allowbreak hierarchy &
data fidelity + smoothness; boundary evidence; topology; stability &
voxel \(\rightarrow\) segment \(\rightarrow\) \newline organ part \(\rightarrow\) organ\\

Microscopy / \newline bioimaging &
voxels/pixels &
morphology; watershed; clustering; SSL representation learning &
partition/\allowbreak hierarchy/\allowbreak field &
contrast; shape priors; persistence; compressibility; robustness &
pixel/voxel \(\rightarrow\) \newline (super)voxel \(\rightarrow\) cell \(\rightarrow\) tissue\\

Seismology / \newline geophysics &
samples / \newline volumes &
coherence filtering; variational surfaces; graph continuity &
field/\allowbreak partition/\allowbreak hierarchy &
signal coherence; discontinuity; cross-scale consistency; stability &
sample \(\rightarrow\) event \(\rightarrow\) \newline horizon \(\rightarrow\) unit\\

Astronomy &
pixels / \newline spatio-spectral samples &
thresholding; clustering; topology/persistence; SSL organisation &
partition/\allowbreak graph/\allowbreak hierarchy &
photometric SNR; coherence; persistence; anomaly/deviation &
pixel \(\rightarrow\) source/blob \(\rightarrow\) object candidate \(\rightarrow\) field\\

Material imaging &
pixels / EBSD grid &
thresholding; morphology; clustering; graph-based &
partition/\allowbreak field/\allowbreak hierarchy &
texture regularity; grain-boundary coherence; deviation from regularity &
pixel \(\rightarrow\) grain segment \(\rightarrow\) phase region \(\rightarrow\) domain\\

Point clouds / \newline 3D sensing &
points &
geometric grouping; graph learning; SSL on geometry &
graph/\allowbreak partition/\allowbreak hierarchy &
geometric consistency; local planarity/curvature; robustness &
point \(\rightarrow\) segment \(\rightarrow\) instance candidate \(\rightarrow\) scene region\\

Robotics &
keypoints/ \newline frames &
bundle adjustment; pose-graph optimisation &
graph (pose graph) &
reprojection consistency; loop closure; stability under drift &
keypoint \(\rightarrow\) landmark \(\rightarrow\) submap \(\rightarrow\) global map\\

\bottomrule
\end{tabular}

\caption{Cross-domain correspondence between carrier sets \(\Omega\), operator types realising \(S_C\), structural-product types, and recurring families of criteria \(C\). Example practice terms illustrate domain usage but are not theory primitives (EBSD = Electron Backscatter Diffraction, SSL = Self-supervised learning, SNR = Signal-to-Noise Ratio).}
\label{tab:crossdomain_scaffold}
\end{sidewaystable}


\section{Structural products as AI-ready FAIR digital objects for digital twins}
\label{sec:structural-products-fair-digital-twins}

A criteria-first, semantics-later approach reshapes what counts as a result and how results should be validated. If the first analytic layer derives a structural product from measurement under explicit criteria, it yields a domain-agnostic, semantics-free structural product \(S\) that is reproducible by design and transferable across tasks and communities. This shift is not an optional add-on. It entails consequences for validation, reproducibility, monitoring under drift, and for research as well as data and data-analytic infrastructure. In particular, criterion-defined structural products can be treated as machine-actionable, versioned digital objects that support FAIR practices \citep{Wilkinson2016,Huerta2023,SoilandReyes2024} and provide stable digital-twin state variables for digital-twin applications \citep{Rossmann2022,NASEM2024DigitalTwins}.

\paragraph{Beyond class accuracy towards structural validation criteria}
If semantics is relocated downstream, validation primarily against semantic \enquote{ground truth} labels becomes untenable as a universal measure of success, especially under open-ended scientific discovery and long-term drift. Evaluation must therefore target the stability and adequacy of the structural product rather than agreement with potentially shifting label sets. A criteria-first approach highlights five evidence classes:
(i) Robustness: stability under perturbations (noise, illumination/sensor drift, seasonality, site variation);
(ii) Scale coherence: consistency across resolutions and scale spaces \citep{Witkin1984,Koenderink1984,Lindeberg2008};
(iii) Complexity control (compressibility): preference for shorter descriptions that preserve salient regularities \citep{Rissanen1978};
(iv) Global optimality and consistency: solutions induced by a well-defined global criterion rather than ad hoc heuristics \citep{Mumford1989,Shi2000}; and
(v) Downstream pluralism: capacity to support multiple semantic mappings on the same \(S\) without hardwiring a single domain ontology upstream.
These evidence classes relocate rigor to reproducible structural adequacy and to explicit, testable stability commitments.

\paragraph{Reproducibility, domain transfer, and open-ended scientific discovery}
Criteria-first makes reproducibility operational. Given \((X,C)\) and a specified implementation, independent teams can reproduce the same structural product \(S\). This also turns domain transfer into a methodological default. The same criteria-first approach can be deployed across modalities and disciplines, with semantic mappings introduced only where domain knowledge is required. For long-term monitoring, the implication is equally direct. Change detection and early warning need not require predefining all novelties as classes. A criterion-defined structural layer supports open-ended scientific discovery by characterising deviations from stable structural regimes in the measurement stream rather than by assigning observations to a fixed label set \citep{Zhu2024,Eyring2024}.

\paragraph{AI-ready, FAIR-by-design, and digital-twin state variables}
Structural products can be treated as first-class research outputs that meet infrastructural requirements. Criteria, software, and workflows should be citable, versioned, and machine-actionable. FAIR has expanded from data to research software and workflows \citep{Barker2022,Huerta2023,Wilkinson2025}, and FAIR Digital Object discussions emphasise persistent identification, consistent metadata, and defined operations for interoperable research outputs \citep{SoilandReyes2024,Peters2024}. A criterion-defined structural product is well suited to this framing. It can be published as a versioned digital object \(\mathrm{DO}=(S,\text{metadata},\text{provenance},\text{version})\). The \(\mathrm{DO}\) metadata should make the criterion, stability commitments, and provenance explicit. At a minimum, reporting a criteria-first structural product should include: (i) the declared criterion \(C\) (objective/constraints) and parameterisation; (ii) an implementation identifier (software version and hash) and random seed, where applicable; (iii) the declared perturbation family or scale envelope used for stability tests; (iv) structural quality metrics (e.g., boundary/partition stability as in Fig.~\ref{fig:one-image-two-criteria}); and (v) an exported stability/uncertainty envelope as part of the \(\mathrm{DO}\) metadata. Operationally, this requires community-maintained schemas or application profiles for \(S\) and its \(\mathrm{DO}\) metadata. It also requires declared operations and lightweight conformance checks so reuse and benchmarking remain comparable across versions and under drift. As versioned, machine-actionable artefacts, such structural products can also serve as standardised inputs or targets for representation learning and foundation-model pipelines, for example as conditioning signals or benchmark substrates. This supports reuse and monitoring under drift without collapsing \(S\) into any single domain ontology. This framing also aligns with AI-readiness approaches, which tie readiness to rigorous documentation, provenance, reliability, and FAIRness \citep{Lawrence2017,Clark2024}. Digital twins intensify these needs: they integrate observations, models, and AI for continuous decision support, yet domain ontologies evolve (e.g., clinical endpoints change, land-cover legends are revised, and taxonomies update) while digital-twin state variables must remain comparable over long horizons \citep{Rossmann2022,NASEM2024DigitalTwins,Tudor2025,Bongomin2025}. A criteria-first approach therefore provides a stable, reproducible, presemantic structural product that can function as a digital twin’s durable state-variable layer. Downstream semantics becomes an interface that can be revised or pluralised without breaking longitudinal comparability \citep{Hazeleger2024}.

\section{Outlook and research agenda}
\label{sec:outlook-agenda-structural}

The proposed approach implies a task-oriented research agenda that is both conceptual and practical:
\begin{itemize}
\item \textbf{Formalise criteria:} identify families of optimality and stability criteria that are scientifically meaningful and computationally tractable across modalities.
\item \textbf{Build structural benchmarks:} evaluate structural products by robustness, scale coherence, compressibility, and global consistency rather than only semantic accuracy.
\item \textbf{Make mappings explicit:} treat interpretation as a documented semantic mapping \(M_i:S\rightarrow \mathcal{O}_i\), enabling transparent pluralism and crosswalks across domain ontologies.
\item \textbf{Use representation learning carefully:} exploit SSL/foundation models as implementation families for criteria-first structure extraction, while keeping criteria explicit and testable \citep{Fotopoulou2024,Li2025}.
\item \textbf{Standardise structural products:} define schemas, versioning, uncertainty/stability envelopes, provenance, and conformance checks for semantics-free structural products as shareable scientific artefacts.
\item \textbf{Separate mapping governance:} treat downstream semantics (semantic mappings and their associated domain ontologies) as its own artefact class so competing mappings can coexist without rewriting upstream structure.
\item \textbf{Build domain-independent tooling:} develop modular, FAIR-aligned software and workflow layers for criteria-first structure extraction \citep{Barker2022,Wilkinson2025}.
\end{itemize}

Recent advances in self-supervised learning strengthen the contact point with this agenda while also making the conceptual gap visible. Fully self-supervised frameworks can learn image representations without category labels, directly probing how much structure can be acquired without top-down domain ontologies \citep{Konkle2022}. Work on aligning machine and human vision likewise suggests that representation quality depends on how distinctions are organised across multiple abstraction levels, implying that questions of structure and level precede downstream naming \citep{Muttenthaler2025}. In label-constrained fields such as medical imaging, reviews document how self-supervised learning is increasingly used to extract transferable structural products from unlabelled data before task-specific semantic mappings are introduced \citep{Huang2023}. The claim here is not that self-supervision automatically delivers semantics-free science, but that it can be interpreted most rigorously when treated as an implementation family for explicit, criteria-first structure extraction.

\section*{Acknowledgements and funding}

This work was supported by the Helmholtz Association, Germany, and the Federal Ministry of Research, Technology and Space (BMFTR), Germany, through the Helmholtz DataHub initiative of the Research Field Earth and Environment. Helmholtz DataHub enables overarching research data management in accordance with the FAIR principles for the Earth and Environment Programme Changing Earth -- Sustaining our Future.\\

This work contributes to the activities of NFDI4Earth and NFDI4BIOIMAGE, funded by the Deutsche Forschungsgemeinschaft (DFG; project number 460036893 for NFDI4Earth and project number 501864659 for NFDI4BIOIMAGE). NFDI4Earth and NFDI4BIOIMAGE are consortia within the National Research Data Infrastructure (NFDI) e.\,V. The NFDI is financed by the Federal Republic of Germany and its 16 federal states.\\

This research also contributes to the development of the eLTER RI cyberinfrastructure within the Horizon 2020 project eLTER PPP (grant no.\ 871126), eLTER PLUS (grant no.\ 871128), and eLTER EnRich (grant no.\ 101131751), funded by the European Commission.\\

\newpage

\appendix

\section{Supplement: Domain overviews}
\label{app:supplementary-note-domain-sketches}

Across image-based sciences, the dominant framing remains semantics-first. Analysis is organised around mapping measurements into a domain ontology or label set \citep{Lv2023,Bom2023,Xu2024,Sun2024,Monteiro2024,Mueller2024}. Yet, in multiple disciplines, criteria-first components recur in practice whenever semantic labelling becomes scarce, unstable, contested, or prohibitively expensive. The recurring pattern is typically hybrid: an upstream, criterion-defined structural product \(S\) is extracted from the measurement stream under explicit criteria, while semantic interpretation remains downstream as a purpose- and community-dependent semantic mapping into a domain ontology/vocabulary. In many current workflows, reporting and benchmarking remain ontology- and label-centric even when a criteria-first sublayer is present.

\paragraph{How to read the sketches}
Each sketch highlights (i) the semantics-first default in the domain (mapping measurements to a domain ontology or label set), (ii) why this becomes limiting under long-term monitoring, domain shift, or open-ended scientific discovery, and (iii) the recurrent criteria-first sublayer(s) that produce transferable structural products (often partitions, graphs, hierarchies, or structural fields) before semantic mappings are applied. The goal is not to claim that the criteria-first inversion is already the dominant paradigm, but to make visible a cross-domain design pattern that is already widely used in components, and to motivate why treating \(S\) as a citable, versioned research artefact is a logical next step.

\subsection*{Earth observation (EO) and environmental monitoring}
\label{subsec:eo-beyond-discretization-supp}

EO products are commonly delivered as thematic discretisations (land-cover/land-use maps, habitat layers, damage classes) whose adequacy is judged by agreement with a chosen domain ontology. In practice, segmentation is often treated as an explicit criterion-driven step that divides images into homogeneous regions, after which classification/labelling acts downstream on the segmented product \citep{Vitti2012}. Under long-term monitoring, semantics-first assumptions become fragile: acquisition conditions and processing chains change, and domain ontologies drift across institutions and policy regimes, making crosswalk-only translations misleading and complicating change analysis \citep{Yang2017}. Upstream comparability is therefore often enforced by explicit criteria on the measurement stream before thematic interpretation, for example via analysis-ready measurement products and reusable spatiotemporal infrastructures such as EO data cubes \citep{Giuliani2019}. Novelty and regime shifts are then first detectable as structural deviations rather than as \enquote{new classes}, and change detection is not reducible to static label agreement \citep{Cheng2024}. Related criteria-first components include object-based image analysis and scale-sensitive landscape structure measures \citep{McGarigal2009,Cushman2010,Blaschke2014}, as well as spectral approaches to biodiversity \citep{Rocchini2021}, aligning with a recent hybrid landscape modelling framework \citep{Lausch2026}. Self-supervised and foundation-model efforts likewise learn transferable structure from large unlabelled archives \citep{Manas2021}. Semantic mappings remain purpose-bound, while reporting and benchmarks remain largely semantics-first; domain shift is typically handled as an add-on to label-centric pipelines rather than as an explicit validation target of the structural product itself \citep{Tuia2016}.

\subsection*{Medical imaging}
\phantomsection\label{subsec:medical-imaging}

Medical imaging workflows commonly centre on supervised or weakly supervised segmentation pipelines that map voxels/pixels to predefined anatomical or pathological label sets, reinforced by benchmarks that focus on agreement with expert-annotated ground truth \citep{Ronneberger2015,Litjens2017,Xu2024}. Label-centric evaluation presumes a stable medical domain ontology, yet clinical vocabularies evolve and labels are expensive, contested, and context-dependent; in addition, inter-rater variability and protocol differences mean that \enquote{ground truth} is often only conditionally well-defined, so comparability and reuse depend less on fixed labels than on reproducible extraction of stable boundaries, regions, and salient structural deviations. Classical segmentation approaches optimise explicit criteria (intensity homogeneity, edge continuity, shape priors) to derive structural products before any semantic mapping is applied \citep{Kass1988,Pham2000}, and this ordering is mirrored in practice when structural anomalies (e.g., lesion boundaries, organ contours) are identified prior to clinical interpretation. Modern workflows also increasingly use self-/weakly supervised representation learning and foundation-style pretraining to extract transferable structure from unlabelled scans before fine-tuning to a specific label set \citep{Zhou2019}. Semantic mappings then attach diagnostic labels or ontology terms and may change as medical ontologies evolve, but in most current pipelines evaluation and reporting still privilege label agreement; intermediate structural products (e.g., boundary fields, instance partitions, uncertainty maps) are not consistently exposed, versioned, and reused as first-class research outputs.

\subsection*{Microscopy / bioimaging}
\phantomsection\label{subsec:microscopy-bioimage}

In bioimaging, especially in modern high-resolution microscopy, the volume and complexity of data make exhaustive manual labelling impractical, yet a common pipeline still attempts semantic segmentation by mapping pixels or voxels to a predefined biological domain ontology (often operationalised as a label set of cell types, subcellular structures, or phenotypes) using supervised learning \citep{Moen2019,Ragone2023}. As modalities such as light-sheet and electron microscopy produce massive data volumes, semantic curation cannot keep pace with data generation; ontology terms for every cell or organelle in a volume may be undefined or too time-consuming to obtain, highlighting the scaling challenge for semantics-first approaches. Microscopy has therefore repeatedly driven criteria-first components whose primary goal is to identify coherent objects or regions without full semantic annotation: deep learning is often first used for representation learning, noise reduction, or generic feature extraction on unlabelled images \citep{Meijering2020}, yielding transferable structural products that support downstream interpretation. Unsupervised segmentation can partition 3D volumes into neuronal segments or organelle regions based on imaging criteria such as membrane continuity or texture \citep{Vincent1991}, long before each segment is mapped into a biological domain ontology; large-scale connectomics underscores the same ordering, where the core bottleneck is reliable extraction of each process from dense imagery \citep{Berning2015}. Only after these structural partitions are obtained do researchers map them to neuronal identities or cell types, and semantic mappings are applied downstream, including via generalist segmentation tools \citep{Carpenter2006,Stringer2021}. Community practice and benchmarks often remain semantics-first: naming, curation, and evaluation dominate what is counted as \enquote{success}, while the structural product is treated as an intermediate step rather than as a stable, portable research output.

\subsection*{Seismology and geophysics}
\phantomsection\label{subsec:seismology-geophysics}

There is a growing trend to apply semantic segmentation approaches to seismic data, for example using deep learning to map regions of a seismic image to a geological domain ontology (e.g., facies or rock types) based on training examples \citep{Bergen2019,Monteiro2024}. Semantics-first pipelines treat seismic interpretation as an image classification task (e.g., sandstone, shale, salt), but traditional -- and in many operational workflows still -- practice emphasises a different ordering: first extract key structural features from the volume, then interpret them geologically, deferring semantic mapping until after partitioning by explicit criteria. Classical workflows focus on picking horizons, discontinuities, and coherent events using signal-processing criteria \citep{Allen1978}; techniques such as semblance analysis, coherence measures, and edge detection identify continuous reflectors and faults based on waveform similarity or discontinuity criteria, and optimisation or threshold criteria guide extraction (e.g., high coherence indicates continuity; low coherence highlights breaks) \citep{Coleou2003}. Modern neural pickers/detectors can likewise be used as criterion-driven extractors \citep{Zhu2018,Perol2018}. The result is a structural product (e.g., reflector and fault surfaces) that can be made stable and reproducible across surveys, while only subsequently are these structures mapped into a stratigraphic or facies ontology (e.g., naming a horizon as the top of a reservoir formation) by an interpreter; many evaluation settings still benchmark progress by agreement with labelled facies maps, so the criteria-first backbone often remains implicit rather than being formalised and shared as a reusable structural product.

\subsection*{Astronomy}
\phantomsection\label{subsec:astronomy}

Astronomical surveys routinely produce label-centric catalogues with semantic annotations, mapping objects into a domain ontology (e.g., galaxy morphology) or distinguishing stars from galaxies \citep{Bertin1996,Lintott2008,Dieleman2015,Bom2023}. At survey scale, exhaustive semantic curation cannot keep pace with the data deluge: predefined domain ontologies inevitably fall short when sources do not fit known categories or exhibit novel combinations of attributes, making astronomy a canonical open-ended scientific discovery setting. Accordingly, in discovery-oriented analyses astronomers increasingly use unsupervised and self-supervised methods to organise survey data and uncover latent structure without immediate reference to established classes \citep{Cheng2021,Fotopoulou2024}. Techniques such as clustering, manifold learning, and representation learning group sources by similarity in a data-driven way, effectively partitioning measurement space into stable regions that initially lack names; self-supervised representations can capture structural features (shapes, brightness profiles, spectral signatures) shared across many objects, and clusters and outliers then become candidates for new phenomena. These data-driven structural products are later mapped to astrophysical interpretations and, where appropriate, to new or revised ontology terms; supervised models can coexist for targeted tasks (e.g., specialised detectors trained on simulated labels) \citep{Hezaveh2017}. Flagship products and evaluation often remain label-centric, so this criteria-first layer—while central in discovery workflows—is rarely exposed as a standardised, versioned structural product.

\subsection*{Materials science}
\phantomsection\label{subsec:materials-science}

In materials imaging and analysis, researchers often begin with a semantics-first framing: segment an image into known phase regions (e.g., crystalline phases, grains) or classify features as particular defect types \citep{Kalinin2015}. Supervised learning often dominates published workflows, mapping microstructural images into a predefined materials domain ontology (often operationalised as a label set of phase labels or defect classes) \citep{DeCost2017,Cecen2018,Mueller2024}. Such semantics-first pipelines assume that all relevant microstructural categories are known in advance and remain stable across imaging conditions, even though boundaries between categories can shift with microscope settings or processing pipelines and relevant phenomena may not fit a fixed label set. Materials science has long employed criteria-first methods to characterise microstructure \citep{Bostanabad2018}: microstructure segmentation often relies on explicit clustering, geometric regularisation, and morphological constraints to derive structural products such as grains, phase boundaries, or anomalous regions without immediate recourse to ontology terms \citep{Kim2020,Mueller2024,Kho2025}. A common example is defect detection, where rather than assuming a universal dictionary of defect types, algorithms can first identify discontinuities or anomalous arrangements by explicit criteria (deviation from expected periodic structure), yielding a reproducible structural product that can be analysed further. Only afterwards are regions mapped to more specific ontology terms (e.g., dislocation, void, second-phase inclusion) based on context, but in much current practice these criteria-first components are embedded within semantics-first pipelines whose reported outputs and benchmarks remain label-centric, and intermediate structural products are seldom curated and shared as reusable, versioned artefacts.

\subsection*{Point clouds and 3D sensing}
\phantomsection\label{subsec:point-clouds-3d}

In 3D scene understanding (e.g., from LiDAR, depth sensors, or multi-view reconstruction), the predominant discourse has been semantic segmentation of point clouds: points are mapped into a label set (e.g., \enquote{ground}, \enquote{building}, \enquote{vehicle}, \enquote{vegetation}), and benchmarks measure success by point-level label accuracy for a fixed ontology \citep{Qi2017,Guo2021,Sun2024}. Semantic labels in 3D data are unstable across environments and sensors: ontology terms shift across settings (e.g., \enquote{building} may not exist in natural scenes), so fixed label sets do not transfer cleanly and require frequent relabelling or adaptation. An alternative emphasis therefore targets geometric segmentation and representation learning that identify stable structural products prior to semantic mapping \citep{Fei2023,Zeng2024}. Many approaches first seek structure in the point cloud itself via geometric criteria (e.g., robust fitting and alignment) \citep{Fischler1981,Besl1992}, grouping points into planar surfaces, geometric clusters, or consistent volumes based on proximity, normals, and multi-scale coherence; the result is a structural product, a decomposition of the point cloud into geometrically coherent parts. Semantic mappings can then attach ontology terms as needed (e.g., map a flat horizontal segment to \enquote{ground}) and can be revised across contexts without recomputing upstream segmentation, but in most published settings success is still primarily measured by semantic label accuracy, so the criteria-first layer often remains a latent backbone rather than an explicitly validated, shareable structural product.

\subsection*{Robotics}
\phantomsection\label{subsec:robotics-slam}

In robotics and autonomous systems, a paradigmatic criteria-first approach is simultaneous localisation and mapping (SLAM) \citep{MurArtal2015,Cadena2016}. Classical visual SLAM builds maps (often sparse point maps or pose graphs) by optimising geometric criteria such as reprojection consistency of features (e.g., keypoints/descriptors combined with robust estimation) \citep{Fischler1981,Lowe2004}, minimising odometry drift, and enforcing loop-closure agreement; the result is a geometric structural product (trajectory and map) without inherent ontology terms. Semantics-first framings arise when tasks are specified as label prediction (including end-to-end semantic prediction from images to actions; \citealp{Bojarski2016}), or when semantic annotations are layered onto maps (\enquote{semantic SLAM}; \citealp{Chen2022}); these additions depend on application-specific ontologies and can change without altering the geometric backbone. The core SLAM process remains a criteria-first structural backbone: a criterion-defined structural product (map and trajectory) is obtained by enforcing geometric consistency constraints and integrating sensor evidence, and this structural product can be made stable and reproducible across runs and platforms when criteria and implementations are fixed. Semantic mappings can then attach ontology terms to map elements (e.g., labelling clusters as \enquote{table} or regions as \enquote{room A}) for decision-making and communication and can be revised without rerunning mapping, but in current research practice semantics is often added as a task layer on top of SLAM without turning the structural backbone into a standardised, benchmarked structural product with explicit stability envelopes.

\newpage

\bibliographystyle{abbrvnat}
\bibliography{260217_criteria-first-semantics-later_arxiv}

\end{document}